# Linear chain conditional random fields, hidden Markov models, and related classifiers

Elie Azeraf [1], Emmanuel Monfrini [2] and Wojciech Pieczynski [2,*]

[1] Watson Department, IBM France, Paris, France; elie.azeraf@ibm.com
[2] Telecom SudParis, Institut Polytechnique de Paris, Evry, France ; {Emmanuel.Monfrini, Wojciech.Pieczynski}@telecom-sudparis.eu
* Correspondence: Wojciech.Pieczynski@telecom-sudparis.eu

**Abstract:** Practitioners use Hidden Markov Models (HMMs) in different problems for about sixty years. Besides, Conditional Random Fields (CRFs) are an alternative to HMMs and appear in the literature as different and somewhat concurrent models. We propose two contributions. First, we show that basic Linear-Chain CRFs (LC-CRFs), considered as different from the HMMs, are in fact equivalent to them in the sense that for each LC-CRF there exists a HMM – that we specify – whom posterior distribution is identical to the given LC-CRF. Second, we show that it is possible to reformulate the generative Bayesian classifiers Maximum Posterior Mode (MPM) and Maximum a Posteriori (MAP) used in HMMs, as discriminative ones. The last point is of importance in many fields, especially in Natural Language Processing (NLP), as it shows that in some situations dropping HMMs in favor of CRFs was not necessary.

## 1. Introduction

Let $Z_{1:N} = (Z_1, \ldots, Z_N)$ be a stochastic sequence, with $Z_n = (X_n, Y_n)$. Random variables $X_1, \ldots, X_N$ take their values in a finite set $\Lambda$, while $Y_1, \ldots, Y_N$ take their values either in a discrete or continuous set $\Omega$. Realizations of $X_{1:N} = (X_1, \ldots, X_N)$ are hidden while realizations of $Y_{1:N} = (Y_1, \ldots, Y_N)$ are observed, and the problem we deal with is to estimate $X_{1:N} = x_{1:N}$ from $Y_{1:N} = y_{1:N}$. We deal with Bayesian methods of estimation, which need some probabilistic model. Probabilistic model is a distribution – or a family of distributions – which will be denoted with $p(z_{1:N})$, or $p(x_{1:N}, y_{1:N})$. We are interested in the case of dependent $Z_1, \ldots, Z_N$. The simplest model taking into account this dependence is the well-known hidden Markov model (HMM) [1, 2, 3, 4, 5], whose distribution is given with

$$p(x_{1:N}, y_{1:N}) = p(x_1)\, p(y_1|x_1) \prod_{n=1}^{N-1} p(x_{n+1}|x_n) p(y_{n+1}|x_{n+1}). \tag{1.1}$$

HMMs allow recursive fast computation of Bayesian estimators called "classifiers" in this paper and recalled below. In spite of their simplicity, HMMs are very robust and provide quite satisfactory results in many applications.

Beside, conditional random fields (CRFs) [6, 7] also allow estimating $X_{1:N} = x_{1:N}$ from $Y_{1:N} = y_{1:N}$. Their definition is different from the definition of HMMs in that in CRFs, one directly considers $p(x_{1:N}|y_{1:N})$, and neither $p(x_{1:N}, y_{1:N})$ nor $p(y_{1:N}|x_{1:N})$ are needed to perform the estimation. In some areas, like Natural Language Processing (NLP), CRFs are preferred over HMCs because $p(x_{1:N}, y_{1:N})$ and $p(y_{1:N}|x_{1:N})$ are difficult to model. General CRFs are written

$$p(x_{1:N}|y_{1:N}) = p(x_1|y_{1:N}) \prod_{n=1}^{N-1} p(x_{n+1}|x_{n:1}, y_{1:N}), \quad (1.2)$$

In this paper we will consider the following basic "linear-chain" CRF (LC-CRF):

$$p(x_{1:N}|y_{1:N}) = \frac{1}{\kappa(y_{1:N})} exp[\sum_{n=1}^{N-1} V_n(x_n, x_{n+1}) + \sum_{n=1}^{N} U_n(x_n, y_n)], \quad (1.3)$$

with $\kappa(y_{1:N})$ the normalizing constant.

Authors usually consider the two families HMMs and CRFs as different [6, 7, 8, 9, 10, 11, 12, 13]. They classify the former in the category of "generative models", while they classify the latter in the category of "discriminative" models.

Considering the simples cases (1.1) and (1.3), we propose two contributions:

(a) We establish an equivalence between HMMs (1.1) and basic linear-chain CRFs (1.3), which completes the results presented in [14].

Let us notice that wanting to compare the two models directly is somewhat misleading. Indeed, HMMs and CRFs are defined with distributions on different spaces. To be precise, we adopt the following definition

**Definition 1**

Let $X_1, \ldots, X_N, Y_1, \ldots, Y_N$ be two stochastic sequences defined above.

(i) we will call "model" a distribution $p(x_{1:N}, y_{1:N})$;

(ii) we will call "conditional model" a distribution $p(x_{1:N}|y_{1:N})$;

(iii) we will say that a model $p(x_{1:N}, y_{1:N})$ is "equivalent" to a conditional model $q(x_{1:N}|y_{1:N})$ if there exists a distribution $r(y_{1:N})$ such that $p(x_{1:N}, y_{1:N}) = q(x_{1:N}|y_{1:N})r(y_{1:N})$;

(iv) we will say that a family of models A is "equivalent" to a family of conditional models B if for each model $p(x_{1:N}, y_{1:N})$ in A there exists an equivalent conditional model $q(x_{1:N}|y_{1:N})$ in B.

According to Definition 1 HMMs are particular "models", while CRFs are particular "conditional models". Then a particular model HMM cannot be equal to a particular conditional model CRF, but it can be equivalent to it.

Our aim is to show that the family of LC-CRFs (1.3) is equivalent to the family of HMMs (1.1). In addition we specify, for each LC-CRF $q(x_{1:N}|y_{1:N})$, a particular HMM $p(x_{1:N}, y_{1:N})$ such that $p(x_{1:N}|y_{1:N}) = q(x_{1:N}|y_{1:N})$.

Let $p(x_{1:N}, y_{1:N}, \theta)$ be an HMM (1.1), with parameters $\theta$. Taking $r(y_{1:N}) = p(y_{1:N}, \theta)$, it is immediate to see that $p(x_{1:N}|y_{1:N}, \theta)$ is an equivalent CRF. The converse is not immediate. Is a given CRF $p(x_{1:N}|y_{1:N}, \theta)$ equivalent to a HMM? If yes, can we find $r(y_{1:N})$ such that $p(x_{1:N}|y_{1:N}, \theta)r(y_{1:N})$ is a HMM? Besides, can we give its (1.1) form? Responding these questions in a simple linear-chain CRF case is our first contribution. More precisely, we show that the family of LC-CRFs (1.3) is equivalent to the family of HMMs (1.1), and we specify, for each LC-CRF $q(x_{1:N}|y_{1:N})$, a particular HMM $p(x_{1:N}, y_{1:N})$ given in the form (1.1), such that $p(x_{1:N}|y_{1:N}) = q(x_{1:N}|y_{1:N})$.

Note that numerous papers addressed similarities between HMMs and linear-chain CRFs [7, 15]; however, to the best of our knowledge, results proposed here, which are mathematically rigorous in the framework of Definition 1, are new.

(b) We show that "generative" estimators MPM and MAP in HMM are computable in a "discriminative" manner, exactly as in LC-CRF.

One of interests of HMMs and CRFs is that in both of them there exist Bayesian classifiers, which allow estimating $x_{1:N}$ from $y_{1:N}$ in a reasonable computer time. As examples, let us consider the "maximum of posterior margins" (MPM) defined with:

$$[g(y_{1:N}) = \hat{x}_{1:N} = (\hat{x}_1, \ldots, \hat{x}_N)] \Leftrightarrow [\forall n = 1, \ldots, N, p(\hat{x}_n|y_{1:N}) = \sup_{x_n}(p(x_n|y_{1:N}))], \quad (1.4)$$

and the "maximum a posteriori" (MAP) is defined with

$$[g(y_{1:N}) = \hat{x}_{1:N}] \Leftrightarrow [p(\hat{x}_{1:N}|y_{1:N}) = \sup_{x_{1:N}}(p(x_{1:N}|y_{1:N}))], \quad (1.5)$$

Note that likely to any other Bayesian classifier, MPM and MAP are independent from $p(y_{1:N})$. This means that in any generative model $p(x_{1:N}, y_{1:N})$, related Bayesian classifier is strictly the same as that related to the equivalent (in the meaning of Definition 1) CRF model $p(x_{1:N}|y_{1:N})$. We see that the distinction between "generative" and "discriminative" classifiers is not justified: all Bayesian classifiers are discriminative. However, in HMM the related MPM and MAP classifiers are computed calling on $p(y_n|x_n)$, while this is not the case in LC-CRF. We show that both MPM and MAP in HMM can also be computed in a "discriminative" way, without calling on $p(y_n|x_n)$. Thus, the use of MPM and MAP in HMM is strictly the same as its use in LC-CRF, which is our second contribution. One of the consequences is that the use of MPMs and MAPs in the two families HMMs and LC-CRFs presents exactly the same interest, in particular in NLP. This shows that abandoning HMMs in favor of LC-CRFs in NLP because of their "generative" nature [6, 7, 8, 9, 16, 17, 18, 19] of related Bayesian classifiers was not justified.

**2. Equivalence between HMMs and simple linear-chain CRFs**

We will use the following Lemma:

**Lemma**

Let $W_{1:N} = (W_1, \ldots, W_N)$ be random sequence, taking its values in a finite set $\Delta$. Then

(i) $W_{1:N}$ is a Markov chain iff there exist $N - 1$ functions $\varphi_1, \ldots, \varphi_{N-1}$ from $\Delta^2$ to $R^+$ such that

$$p(w_1, \ldots, w_N) \propto \varphi_1(w_1, w_2) \ldots \varphi_{N-1}(w_{N-1}, w_N), \tag{2.1}$$

where "$\propto$" means "proportional to";

(ii) for HMM defined with $\varphi_1, \ldots, \varphi_{N-1}$ verifying (2.1), $p(w_1)$ and $p(w_{n+1}|w_n)$ are given with

$$p(w_1) = \frac{\beta_1(w_1)}{\sum_{w_1} \beta_1(w_1)} \ ; \ p(w_{n+1}|w_n) = \frac{\varphi_n(w_n, w_{n+1})\beta_{n+1}(w_{n+1})}{\beta_n(w_n)}, \tag{2.2}$$

where $\beta_1(w_1), \ldots, \beta_N(w_N)$ are defined with the following backward recursion:

$$\beta_N(w_N)=1, \ \beta_n(w_n) = \sum_{w_{n+1}} \varphi_n(w_n, w_{n+1})\beta_{n+1}(w_{n+1}) \tag{2.3}$$

**Proof of Lemma.**
1. Let $W_{1:N}$ be Markov: $p(w_1, \ldots, w_N) = p(w_1)p(w_2|w_1)p(w_3|w_2) \ldots p(w_N|w_{N-1})$. Then (2.1) is verified by $\varphi_1(w_1, w_2) = p(w_1)p(w_2|w_1)$, $\varphi_2(w_2, w_3) = p(w_3|w_2)$, …, $\varphi_{N-1}(w_{N-1}, w_N) = p(w_N|w_{N-1})$.

2. Conversely, let $p(w_1, \ldots, w_N)$ verifies (2.1). Thus $p(w_1, \ldots, w_N) = K\varphi_1(w_1, w_2) \ldots \varphi_{N-1}(w_{N-1}, w_N)$ with $K$ constant. This implies that for each $n = 1, \ldots, N - 1$ we have

$$p(w_{n+1}|w_1, \ldots, w_n) = \frac{p(w_1, \ldots, w_n, w_{n+1})}{p(w_1, \ldots, w_n)} =$$

$$\frac{\sum_{(w_{n+2}, \ldots, w_N)} \varphi_1(w_1, w_2) \ldots \varphi_n(w_n, w_{n+1})\varphi_{n+1}(w_{n+1}, w_{n+2}) \ldots \varphi_{N-1}(w_{N-1}, w_N)}{\sum_{(w_{n+1}, w_{n+2}, \ldots, w_N)} \varphi_1(w_1, w_2) \ldots \varphi_n(w_n, w_{n+1})\varphi_{n+1}(w_{n+1}, w_{n+2}) \ldots \varphi_{N-1}(w_{N-1}, w_N)} = \tag{2.4}$$

$$\frac{\varphi_n(w_n, w_{n+1}) \sum_{(w_{n+2}, \ldots, w_N)} \varphi_{n+1}(w_{n+1}, w_{n+2}) \ldots \varphi_{N-1}(w_{N-1}, w_N)}{\sum_{(w_{n+1}, w_{n+2}, \ldots, w_N)} \varphi_n(w_n, w_{n+1})\varphi_{n+1}(w_{n+1}, w_{n+2}) \ldots \varphi_{N-1}(w_{N-1}, w_N)} = p(w_{n+1}|w_n),$$

which shows that $p(w_1, \ldots, w_N)$ is Markov.
Besides, let us set $\beta_n(w_n) = \sum_{(w_{n+1}, w_{n+2}, \ldots, w_N)} \varphi_n(w_n, w_{n+1}) \ldots \varphi_{N-1}(w_{N-1}, w_N)$ for $n = 1, \ldots, N - 1$. On the one hand, we see that $\beta_n(w_n) = \sum_{w_{n+1}} \varphi_n(w_n, w_{n+1})\beta_{n+1}(w_{n+1})$. On the other hand, according to (2.4) we have $p(w_{n+1}|w_n) = \frac{\varphi_n(w_n, w_{n+1})\beta_{n+1}(w_{n+1})}{\beta_n(w_n)}$. As $p(w_1) = \frac{\beta_1(w_1)}{\sum_{w_1} \beta_1(w_1)}$, (2.2) and (2.3) are verified, which ends the proof □

Proposition 1 below shows that "linear-chain" CRF defined with (1.3) is equivalent to a HMM defined with (1.1). In addition, $p(x_1)$, $p(x_{n+1}|x_n)$, and $p(y_n|x_n)$ in (1.1) defining an equivalent HMM are computed from $V_n(x_n, x_{n+1})$ and $U_n(x_n, y_n)$. To the best of our knowledge, except some first weaker results in [15], these results are new.

**Proposition 1.** *Let $Z_{1:N} = (Z_1, ..., Z_N)$ be stochastic sequence, with $Z_n = (X_n, Y_n)$. Each $(X_n, Y_n)$ takes its values in $\Lambda \times \Omega$, with $\Lambda$ and $\Omega$ finite. If $Z_{1:N}$ is a linear chain conditional random field (LC-CRF) with the distribution $p(x_{1:N}|y_{1:N})$ defined by*

$$p(x_{1:N}|y_{1:N}) = \frac{1}{\kappa(y_{1:N})} exp[\sum_{n=1}^{N-1} V_n(x_n, x_{n+1}) + \sum_{n=1}^{N} U_n(x_n, y_n)]. \quad (2.5)$$

*then (2.5) is the posterior distribution of the HMM*

$$q(x_{1:N}, y_{1:N}) = q_1(x_1)q(y_1|x_1) \prod_{n=1}^{N-1} q_{n+1}(x_{n+1}|x_n)q(y_{n+1}|x_{n+1}), \quad (2.6)$$

*with*

$$q(x_1) = \frac{\sum_{y_1} \gamma_1(x_1, y_1)}{\sum_{(x_1, y_1)} \gamma_1(x_1, y_1)}; \quad q(y_1|x_1) = \frac{\gamma_1(x_1, y_1)}{\sum_{y_1} \gamma_1(x_1, y_1)} \quad (2.7)$$

$$q(x_{n+1}|x_n) = \frac{\psi(x_{n+1}) exp\,[V_n(x_n, x_{n+1})]}{\sum_{x_{n+1}} \psi(x_{n+1}) exp\,[V_n(x_n, x_{n+1})]}; \quad (2.8)$$

$$q(y_{n+1}|x_{n+1}) = \frac{exp\,[U_{n+1}(x_{n+1}, y_{n+1})] \gamma_{n+1}(x_{n+1}, y_{n+1})}{\psi(x_{n+1})}, \quad (2.9)$$

*where*

$$\psi(x_{n+1}) = \sum_{y_{n+1}} exp[U(x_{n+1}, y_{n+1})]\gamma_{n+1}(x_{n+1}, y_{n+1}), \quad (2.10)$$

*and $\gamma_1(x_1, y_1), ..., \gamma_N(x_N, y_N)$ are given by the backward recursion*

$\gamma_N(x_N, y_N) = 1$, *for $n = N-1, ..., 2$*

$$\gamma_n(x_n, y_n) = \sum_{(x_{n+1}, y_{n+1})} exp\,[V_n(x_n, x_{n+1}) + U_{n+1}(x_{n+1}, y_{n+1})]\gamma_{n+1}(x_{n+1}, y_{n+1}), \quad (2.11)$$

*and $\gamma_1(x_1, y_1) = \sum_{(x_2, y_2)} exp[V_1(x_1, x_2) + U_1(x_1, y_1) + U_2(x_2, y_2)]\gamma_2(x_2, y_2)$.*

**Proof of Proposition 1.** According to the Lemma, functions $\varphi_1, .., \varphi_N$ defined on $\Delta^2$, with $\Delta = \Lambda \times \Omega$, by $\varphi_1(x_1, y_1, x_2, y_2) = exp[V_1(x_1, x_2) + U_1(x_1, y_1) + U_2(x_2, y_2)]$, and for $n = 2, ..., N-1$, $\varphi_n(x_n, y_n, x_{n+1}, y_{n+1}) = exp\,[V_n(x_n, x_{n+1}) + U_{n+1}(x_{n+1}, y_{n+1})]$ define a Markov chain $Z_{1:N} = (Z_1, ..., Z_N)$, with $Z_n = (X_n, Y_n)$. Let us denote with $r(z_{1:N}) = r(x_{1:N}, y_{1:N})$ its distribution. As $r(x_{1:N}, y_{1:N}) = K exp[\sum_{n=1}^{N-1} V_n(x_n, x_{n+1}) + \sum_{n=1}^{N} U_n(x_n, y_n)]$ with $K$ constant, we have $r(x_{1:N}|y_{1:N}) = p(x_{1:N}|y_{1:N})$. Let us show that $r(x_{1:N}, y_{1:N})$ verifies (2.6)-(2.11). According to the lemma, for $n = N-1, ..., 1$:

$$r(x_{n+1}, y_{n+1}|x_n, y_n) = \frac{\varphi_n(x_n, y_n, x_{n+1}, y_{n+1})\gamma_{n+1}(x_{n+1}, y_{n+1})}{\sum_{(x_{n+1}, y_{n+1})} \varphi_n(x_n, y_n, x_{n+1}, y_{n+1})\gamma_{n+1}(x_{n+1}, y_{n+1})} =$$
$$\frac{exp\,[V_n(x_n, x_{n+1}) + U_{n+1}(x_{n+1}, y_{n+1})]\gamma_{n+1}(x_{n+1}, y_{n+1})}{\sum_{(x_{n+1}, y_{n+1})} exp\,[V_n(x_n, x_{n+1}) + U_{n+1}(x_{n+1}, y_{n+1})]\gamma_{n+1}(x_{n+1}, y_{n+1})} \quad (2.12)$$

Introducing $\psi(x_{n+1})$ defined with (2.10), (2.12) writes

$r(x_{n+1}, y_{n+1}|x_n, y_n) = \frac{exp\,[V_n(x_n, x_{n+1})]exp\,[U_{n+1}(x_{n+1}, y_{n+1})]\gamma_{n+1}(x_{n+1}, y_{n+1})}{\sum_{x_{n+1}} \psi(x_{n+1})exp\,[V_n(x_n, x_{n+1})]} =$

$\frac{\psi(x_{n+1})exp\,[V_n(x_n, x_{n+1})]exp\,[U_{n+1}(x_{n+1}, y_{n+1})]\gamma_{n+1}(x_{n+1}, y_{n+1})}{\psi(x_{n+1})\sum_{x_{n+1}} \psi(x_{n+1})exp\,[V_n(x_n, x_{n+1})]} =$

$\left[\frac{\psi(x_{n+1})exp\,[V_n(x_n, x_{n+1})]}{\sum_{x_{n+1}} \psi(x_{n+1})exp\,[V_n(x_n, x_{n+1})]}\right]\left[\frac{exp\,[U_{n+1}(x_{n+1}, y_{n+1})]\gamma_{n+1}(x_{n+1}, y_{n+1})}{\psi(x_{n+1})}\right] =$

$r(x_{n+1}|x_n)r(y_{n+1}|x_{n+1})$,

with $r(x_{n+1}|x_n)$, $r(y_{n+1}|x_{n+1})$ verifying (2.8) and (2.9), respectively.

Beside, $r(x_1, y_1) = \frac{\gamma_1(x_1, y_1)}{\sum_{(x_1, y_1)} \gamma_1(x_1, y_1)}$, which gives (2.7).

Finally, $r(x_{1:N}, y_{1:N})$ verifies (2.6)-(2.11), ends the proof □

## 3. Discriminative classifiers in generative HMMs

One of interests of HMMs and some CRFs with hidden discrete finite data lies in possibilities of analytic fast computation of Bayesian classifiers. As examples of classic Bayesian classifiers, let us consider the MPM (1.4) and the MAP (1.5). However, in some domains like NLP, CRFs are preferred to HMMs for the following reasons.

As HMM is a generative model, MPM and MAP used in HMM are also called "generative", and people consider that HMM based MPM and MAP need the knowledge of $p(y_n|x_n)$. Then people consider it as improper to use them in situations where these distributions are hard to handle. We show that this reason is not valid. More precisely, we show two points:

(i) First, we notice that whatever distribution $p(x_{1:N}, y_{1:N})$, all Bayesian classifiers are independent from $p(y_{1:N})$, so that the distinction between « generative » and « discriminative » classifiers is misleading: they are all discriminative;

(ii) Second, "discriminative" computation of MPM and MAP in HMMs is not intrinsic to HMMs but is due to its particular classic parameterization (1.1). In other words, changing the parametrization, it is possible to compute the HMM based MPM and MAP calling neither on $p(y_{1:N}|x_{1:N})$ nor of $p(y_{1:N})$.

The first point is rather immediate: we note that Bayesian classifier $g_L$ is defined by a loss function $L: \Omega^2 \to \mathbb{R}^+$ through

$$[g_L(y_{1:N}) = \hat{x}_{1:N}] \Leftrightarrow [E[L(g_L(y_{1:N}), X_{1:N})|y_{1:N}] = \inf_{x_{1:N}} E[L(x_{1:N}, X_{1:N})|y_{1:N}]; \quad (3.1)$$

it is thus immediate to notice that $g_L(y_{1:N})$ only depends on $p(x_{1:N}|y_{1:N})$. This implies that it is the same in a generative model or its equivalent (within the meaning of Definition 1.1) discriminative model.

We show (ii) by separately considering MPM and MAP cases.

### 3.1. Discriminative computing of HMM based MPM

To show (ii) let us consider (1.1) with $p(y_n|x_n) = \frac{p(y_n)p(x_n|y_n)}{p(x_n)}$. It becomes

$$p(x_{1:N}, y_{1:N}) = p(x_1|y_1) \prod_{t=2}^{T} p(x_n|x_{n-1}) \frac{p(x_n|y_n)}{p(x_n)} \prod_{n=1}^{N} p(y_n), \quad (3.2)$$

We see that (3.2) is of the form $p(x_{1:N}, y_{1:N}) = h(x_{1:N}, y_{1:N}) \prod_{n=1}^{N} p(y_n)$, where $h(x_{1:N}, y_{1:N})$ does not depend on $p(y_1), \ldots, p(y_N)$. This implies that $h(x_n, y_{1:N}) = \sum_{(x_{1:n-1}, x_{n+1:N})} h(x_{1:N}, y_{1:N})$ does not depend on $p(y_1), \ldots, p(y_N)$ either. Then $p(x_n|y_{1:N}) = \frac{p(x_n, y_{1:N})}{p(y_{1:N})} = \frac{h(x_n, y_{1:N}) \prod_{n=1}^{N} p(y_n)}{[\sum_{x_n} h(x_n, y_{1:N})] \prod_{n=1}^{N} p(y_n)} = \frac{h(x_n, y_{1:N})}{[\sum_{x_n} h(x_n, y_{1:N})]}$, so that

$$p(x_n|y_{1:N}) = \frac{\sum_{(x_{1:n-1}, x_{n+1:N})} p(x_1|y_1) \prod_{t=2}^{N} p(x_n|x_{n-1}) \frac{p(x_n|y_n)}{p(x_n)}}{\sum_{(x_{1:N})} p(x_1|y_1) \prod_{t=2}^{N} p(x_n|x_{n-1}) \frac{p(x_n|y_n)}{p(x_n)}} \quad (3.3)$$

neither depends on $p(y_1), \ldots, p(y_N)$. Thus, HMM based classifier MPM also verifies the "discriminative classifier" definition.

How to compute $p(x_n|y_{1:N})$? It is classically computable using "forward" probabilities $\alpha_n(x_n)$ and "backward" ones $\beta_n(x_n)$ defined with

$$\alpha_n(x_n) = p(x_n, y_{1:n}), \quad (3.4)$$

$$\beta_n(x_n) = p(y_{n+1:N}|x_n). \quad (3.5)$$

Then

$$p(x_n|y_{1:N}) = \frac{\alpha_n(x_n)\beta_n(x_n)}{\sum_{x_n} \alpha_n(x_n)\beta_n(x_n)}, \quad (3.6)$$

with all $\alpha_n(x_n)$ and $\beta_n(x_n)$ computed using the following forward and backward recursions [20]:

$$\alpha_1(x_1) = p(x_1)p(y_1|x_1); \ \alpha_{n+1}(x_{n+1}) = \sum_{x_n} p(x_{n+1}|x_n)p(y_{n+1}|x_{n+1})\alpha_n(x_n), \quad (3.7)$$

$$\beta_N(x_N) = 1; \ \beta_n(x_n) = \sum_{x_{n+1}} p(x_{n+1}|x_n)p(y_{n+1}|x_{n+1})\beta_{n+1}(x_{n+1}). \quad (3.8)$$

Setting $p(y_n|x_n) = \frac{p(y_n)p(x_n|y_n)}{p(x_n)}$ and recalling that $p(x_n|y_{1:N})$ does not depend on $p(y_1)$, …, $p(y_N)$, we can arbitrarily modify them. Let us consider the uniform distribution over $\Omega$, so that $p(y_1) = \cdots = p(y_N) = \frac{1}{\#\Omega} = c$. Then (3.7), and (3.8) become

$$\alpha^*_{n+1}(x_{n+1}) = \sum_{x_n} p(x_{n+1}|x_n)\frac{cp(x_{n+1}|y_{n+1})}{p(x_{n+1})}\alpha^*_n(x_n) \quad ; \quad (3.9)$$

$$\beta^*_n(x_n) = \sum_{x_{n+1}} p(x_{n+1}|x_n)\frac{cp(x_{n+1}|y_{n+1})}{p(x_{n+1})}\beta^*_{n+1}(x_{n+1}), \quad (3.10)$$

and we still have

$$p(x_n|y_{1:N}) = \frac{\alpha^*_n(x_n)\beta^*_n(x_n)}{\sum_{x_n}\alpha^*_n(x_n)\beta^*_n(x_n)}. \quad (3.11)$$

Finally, we see that $p(x_n|y_{1:N})$ is independent from $c$, so that we can take $c = 1$. Then we can state

**Proposition 2** *Let $X_1, …, X_N, Y_1, …, Y_N$ be a HMM (1.1). Let define "forward discriminative" quantities $\alpha^D_1(x_1), …, \alpha^D_N(x_N)$, and "backward discriminative" ones $\beta^D_1(x_1), …, \beta^D_N(x_N)$ by the following forward and backward recursions:*

$$\alpha^D_1(x_1) = p(x_1|y_1); \ \alpha^D_{n+1}(x_{n+1}) = \sum_{x_n} p(x_{n+1}|x_n)\frac{p(x_{n+1}|y_{n+1})}{p(x_{n+1})}\alpha^D_n(x_n), \quad (3.12)$$

$$\beta^D_N(x_N) = 1; \ \beta^D_n(x_n) = \sum_{x_{n+1}} p(x_{n+1}|x_n)\frac{p(x_{n+1}|y_{n+1})}{p(x_{n+1})}\beta^D_{n+1}(x_{n+1}). \quad (3.13)$$

*Then*

$$p(x_n|y_{1:N}) = \frac{\alpha^D_n(x_n)\beta^D_n(x_n)}{\sum_{x_n}\alpha^D_n(x_n)\beta^D_n(x_n)}. \quad (3.14)$$

*Consequently, we can compute MPM classifier in a discriminative manner, only using $p(x_1), …, p(x_N)$, $p(x_2|x_1), …, p(x_N|x_{N-1})$, and $p(x_1|y_1), …, p(x_N|y_N)$.*

Then for $n = 1, …, N$, $p(x_n|y_{1:N})$ can be computed par the only use of $p(x_1), …, p(x_N)$, $p(x_2|x_1), …, p(x_N|x_{N-1})$, and $p(x_1|y_1), …, p(x_N|y_N)$. This shows that $p(x_n|y_{1:N})$ can be computed with the only use of $p(x_n|y_n)$ and $p(x_n)$, exactly as in CRF case. For example, in supervised NLP problems $p(x_n|y_n)$ and $p(x_n)$ can be easily estimated and thus the interest of HMMs is equivalent to that of CRFs. This result is similar to [21], but with a more concise proof.

**Remark 1**

Let us notice that according to (3.12), (3.13), it is possible to compute $\alpha^D_n(x_n)$ and $\beta^D_n(x_n)$ by a very slight adaptation of classic computing programs giving classic $\alpha_n(x_n) = p(x_n, y_{1:n})$ and $\beta_n(x_n) = p(y_{n+1:N}|x_n)$ with recursions (3.7), (3.8). All we have to do is to replace $p(y_{n+1}|x_{n+1})$ with $\frac{p(x_{n+1}|y_{n+1})}{p(x_{n+1})}$. Of course, $\alpha^D_n(x_n) \neq p(x_n, y_{1:n})$ and $\beta^D_n(x_n) \neq p(y_{n+1:N}|x_n)$, but (3.14) holds. This is the core point: one uses a wrong model to find $p(x_n|y_{1:N})$ of the true HMC, that is the final goal.

**Remark 2**

We see that we can compute MPM in HMM only using $p(x_1), …, p(x_N)$ and $p(x_1|y_1), …, p(x_N|y_N)$. This means that in supervised classification, where we have a learn sample, we can use any parametrization to estimate them. For example, we can model them with logistic regression, as currently do in CRFs. It of importance to note that such a parametrization is unusual; however, important is that the model remains the same.

*3.2. Discriminative computing of HMM based MAP: discriminative Viterbi*

Let $X_1, ..., X_N, Y_1, ..., Y_N$ be a HMM (1.1). Bayesian MAP classifier (1.5) based on is computed with the following Viterbi algorithm [22]. For each $n = 1, ..., N$, and each $x_n$, let $x_{1:n-1}^{max}(x_n) = (x_1^{max}, ..., x_{n-1}^{max})(x_n)$ be the path $x_1^{max}, ..., x_{n-1}^{max}$ verifying

$$p(x_{1:n-1}^{max}(x_n), x_n, y_{1:n}) = \sup_{x_{1:n-1}} p(x_{1:n-1}, x_n, y_{1:n}); \qquad (3.15)$$

We see that $x_{1:n-1}^{max}(x_n)$ is a path maximizing $p(x_{1:n-1}, x_n | y_{1:n})$ over all paths ending in $x_n$. Then having the paths $x_{1:n-1}^{max}(x_n)$ and the probabilities $p(x_{1:n-1}^{max}(x_n), x_n, y_{1:n})$ for each $x_n$, one determines, for each $x_{n+1}$, the paths $x_{1:n}^{max}(x_{n+1})$ and the probabilities $p(x_{1:n}^{max}(x_{n+1}), x_{n+1}, y_{1:n+1}) = p(x_{1:n-1}^{max}(x_n^{max}), x_n^{max}, x_{n+1}, y_{1:n+1})$, searching $x_n^{max}$ with

$$p(x_{1:n-1}^{max}(x_n^{max}), x_n^{max}, x_{n+1}, y_{1:n+1})) = \qquad (3.16)$$

$$\sup_{x_n}[p(x_{1:n-1}^{max}(x_n), x_n, y_{1:n})p(x_{n+1}|x_n)p(y_{n+1}|x_{n+1})].$$

Setting in (3.16) $p(y_{n+1}|x_{n+1}) = \frac{p(y_{n+1})p(x_{n+1}|y_{n+1})}{p(x_{n+1})}$, we see that $x_n^{max}$ which verifies (3.16) is the same that $x_n^{max}$ which maximizes $p(x_{1:n-1}^{max}(x_n), x_n, y_{1:n})p(x_{n+1}|x_n)\frac{p(x_{n+1}|y_{n+1})}{p(x_{n+1})}$, so that we can suppress $p(y_{n+1})$. In other words we can replace (3.16) with

$$p(x_{1:n-1}^{max}(x_n^{max}), x_n^{max}, x_{n+1}, y_{1:n+1})) = \qquad (3.17)$$

$$\sup_{x_n}[p(x_{1:n-1}^{max}(x_n), x_n, y_{1:n})p(x_{n+1}|x_n)\frac{p(x_{n+1}|y_{n+1})}{p(x_{n+1})}]$$

Finally, we propose the following discriminative version of the Viterbi algorithm:

- set $x_1^{max} = \underset{x_n}{argmax}[p(x_1|y_1)]$;
- for each $n = 1, ..., N-1$, and each $x_{n+1}$, apply (3.17) to find a path $x_{1:n}^{max}(x_{n+1})$ from the paths $x_{1:n-1}^{max}(x_n)$ (for all $x_n$), and the probabilities $p(x_{1:n}^{max}(x_{n+1}), x_{n+1}, y_{1:n+1})$ (for all $x_{n+1}$);
- end setting $x_{1:N}^{max} = \underset{x_N}{argmax}[p(x_{1:N-1}^{max}(x_N), x_N, y_{1:N})]$.

Likely for MPM above, we see that we can find $x_{1:N}^{max}$ with the only use of $p(x_1), ..., p(x_N), p(x_2|x_1), ..., p(x_N|x_{N-1})$, and $p(x_1|y_1), ..., p(x_N|y_N)$, exactly as in CRF case. As above, it appears that dropping HMMs in some NLP tasks on the grounds that MAP is a "generative" classifier, is not justified. In particular, in supervised stationary framework distributions $p(x_n)$, $p(x_{n+1}|x_n)$, and $p(x_n|y_n)$ can be estimated in the same way as in LC-CRFs case.

## 4. Discussion

We proposed two results. We showed that the basic LC-CRF (1.3) is equivalent to a classic HMM (1.1) in that one can find a HMM whose posterior distribution is exactly the given LC-CRF. More precisely, we specified the way of computing $p(x_1)$ and $p(x_{n+1}|x_n)$, $p(y_n|x_n)$ defining (1.1) from $V_n(x_n, x_{n+1})$, $U_n(x_n, y_n)$ defining (1.3). Then, noticing that all Bayesian classifiers are discriminative in that they do not depend on the observations distribution, we showed that HMMs based classifiers, usually considered as "generative" ones, can also be considered as discriminative classifiers. More precisely, we proposed discriminative ways of computing the classic HMMs based Maximum Posterior Mode (MPM) and Maximum a Posteriori (MAP) classifiers.

We considered basic LC-CRFs and HMMs; extending the proposed results to more sophisticated models like Pairwise Markov chains" (PMCs [23]) is an interesting perspective for further investigations.

**Author Contributions:** Conceptualization, E.A., E.M. and W.P.; methodology, E.A., E.M. and W.P.; validation, E.A., E.M. and W.P.; investigation, E.A., E.M. and W.P.; writing—original draft preparation, W.P.; writing—review and editing, E.A., E.M. and W.P; supervision, W.P.; project administration, W.P. All authors have read and agreed to the published version of the manuscript.

**Funding:** This research was partly funded by the French Government Agency ASSOCIATION NATIONALE RECHERCHE TECHNOLOGIES (ANRT).

**Conflicts of Interest:** The authors declare no conflict of interest.